\documentclass{article}
\usepackage{spconf,amsmath,graphicx}

\usepackage{caption}
\usepackage{subcaption}
\usepackage{spconf,amsmath,epsfig}
\usepackage{amsmath}
\usepackage{epsfig}
\usepackage{graphicx}
\usepackage{algorithm2e}
\usepackage{amssymb}
\usepackage{booktabs}
\usepackage{tabularx}
\usepackage{hyperref}
\hypersetup{
    colorlinks=true,
    linkcolor=black,
    citecolor=black,
    urlcolor=blue,
}
\usepackage{multirow}

\usepackage{cite}

\newcommand\blfootnote[1]{%
  \begingroup
  \renewcommand\thefootnote{}\footnote{#1}%
  \addtocounter{footnote}{-1}%
  \endgroup
}

\title{An Unsupervised Cross-Modal Hashing Method Robust to Noisy Training Image-Text Correspondences in Remote Sensing}
\name{Georgii~Mikriukov, Mahdyar~Ravanbakhsh, Begüm~Demir}
\address{Technische Universität Berlin, Berlin, Germany}
\begin{document}
\ninept
\maketitle
\begin{abstract}

The development of accurate and scalable cross-modal image-text retrieval methods, where queries from one modality (e.g., text) can be matched to archive entries from another (e.g., remote sensing image) has attracted great attention in remote sensing (RS). Most of the existing methods assume that a reliable multi-modal training set with accurately matched text-image pairs is existing. However, this assumption may not always hold since the multi-modal training sets may include noisy pairs (i.e., textual descriptions/captions associated to training images can be noisy), distorting the learning process of the retrieval methods. To address this problem, we propose a novel unsupervised cross-modal hashing method robust to the noisy image-text correspondences (CHNR). CHNR consists of three modules: 1) feature extraction module, which extracts feature representations of image-text pairs; 2) noise detection module, which detects potential noisy correspondences; and 3) hashing module that generates cross-modal binary hash codes. The proposed CHNR includes two training phases: i) meta-learning phase that uses a small portion of clean (i.e., reliable) data to train the noise detection module in an adversarial fashion; and ii) the main training phase for which the trained noise detection module is used to identify noisy correspondences while the hashing module is trained on the noisy multi-modal training set. Experimental results show that the proposed CHNR outperforms state-of-the-art methods. \blfootnote{Our code is publicly available at \url{https://git.tu-berlin.de/rsim/chnr}}

\end{abstract}
%
\section{Introduction}
\label{sec:intro}

The fast-growing volume of multi-modal data (e.g., satellite images acquired by different sensors and their textual descriptions) archives in remote sensing (RS) has attracted great attention for the development of cross-modal retrieval methods. Cross-modal retrieval (CMR), given a query from one modality, aims at retrieving its counterpart from another modality. Among CMR tasks in RS, the image–text retrieval is one of the most challenging tasks because of the huge differences between the representations of RS image and text modalities. The existing cross-modal image-text retrieval methods in RS are defined based on supervised retrieval algorithms, which require the availability of a multi-modal training set with accurately matched text-image pairs. The quantity and the quality of the available image-text training pairs are crucial for achieving accurate cross-modal retrieval. However, collecting a sufficient number of reliable pairs is time-consuming and costly. Unlike RS, in the computer vision (CV) community, unsupervised and self-supervised cross-modal representation learning methods (which rely on the accurate matching between the modalities) are widely studied \cite{jung2021contrastive,su2019deep,liu2020joint, chen2020improved,chen2020simple,he2020momentum}. Su et al. \cite{su2019deep} introduce a deep joint-semantics reconstructing hashing (DJSRH) method to learn binary codes that preserve the neighborhood structure in the original data. To this end, DJSRH learns a mapping from different modalities into a joint-semantics affinity matrix. The use of hashing allows mapping high-dimensional feature vectors into compact binary hash codes, which are indexed into a hash table that enables scalable search and retrieval. Liu et al. \cite{liu2020joint} propose a joint-modal distribution-based similarity weighting (JDSH) method based on DJSRH, exploiting an additional objective based on cross-modal semantic similarities among samples. Unsupervised contrastive learning, which aims to learn a metric space using the sample augmentations (different views of the sample) is introduced in \cite{chen2020simple}. The feature space is learned in a way, that views of one sample are pulled closer together and further from views of other samples from the batch. The existing unsupervised contrastive learning methods mainly rely on inter-modality contrastive objectives to obtain consistent representations across different modalities, while the intra-modal contrastive objectives are ignored. This may lead to learning an inefficient embedding space, where the same semantic content can be mapped into different points in the embedding space \cite{zolfaghari2021crossclr}. The success of the above-mentioned methods also depends on the assumption that the multi-modal training data are correctly matched between modalities (e.g., each training image is associated with an accurate text sentence). However, manually collecting such accurate training sets is costly and time-consuming, and the multi-modal correspondences can be noisy (i.e., the text may not describe the corresponding image content accurately), leading to a training set that includes noisy correspondences. In detail, several factors can introduce noise in the captions. For example, in automatic model-generated captions, noise can occur due to noisy class labels assigned to the data. Manually-generated captions through crowd-sourcing could be subject to noisy captions due to human errors and/or subjectivity in describing the image content. In general, noise in the captions can be due to: 1) the wrong (foil) words \cite{shekhar2017foil}, where one or several words in the caption may not be related to the image content; 2) the missing words, when the caption does not represent all land-use and land-cover class presented in the image; 3) the miscaptioning (wrong caption), where the caption is semantically correct but does not correspond the image; 4) the redundancy \cite{zhang2019denoising, qin2021tvdim}, where the image description contains too much redundant information; 5) the typos and spell-check errors \cite{malykh2019robust}. These issues lead to the construction of a cross-modal training set with noisy image-text correspondences, which may drastically reduce the CMR performance. To address this problem, in this paper we introduce a novel unsupervised cross-modal hashing method robust to the noisy image-text correspondences (CHNR). The proposed CHNR: i) identifies noisy image-text pairs; ii) considers intra- and inter-modal objectives for cross-modal representation learning; and iii) allows high time-efficient search capability. 

\begin{figure*}[t]
  \centering
  \includegraphics[width=0.85\linewidth]{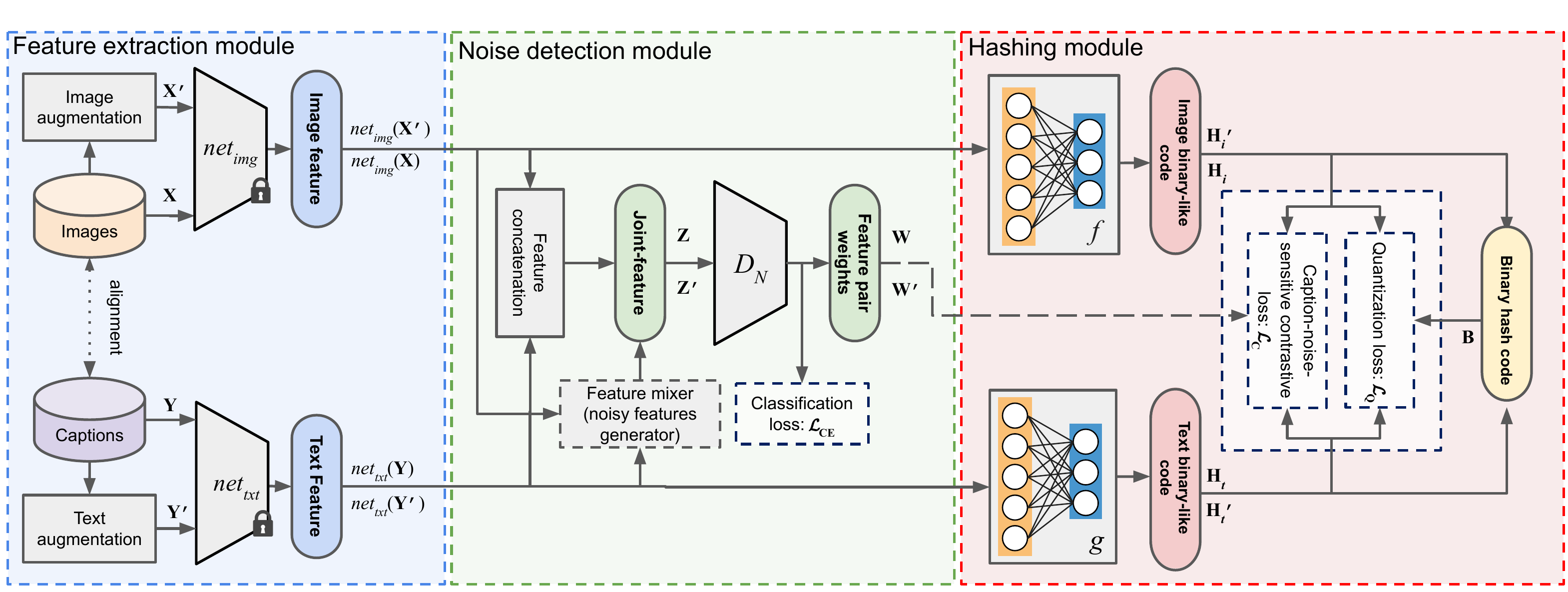}
  \caption{
Block diagram of the proposed CHNR. In the feature extraction module deep feature representations are extracted with the modality-specific encoders $net_{img}$ and $net_{txt}$ for image and text modalities, respectively. The discriminator $D_N$ of the noise detection module detects potential noisy correspondences. The hashing module learns two hash functions $f$ and $g$ from the input embedding.}
  \label{fig:main-diagram}
\end{figure*}




\section{Proposed Method}
\label{sec:method}

Let $\textbf{O} = \{\textbf{X}, \textbf{Y}\}^N$ be a multi-modal training set of $N$ image-text pairs, where $\textbf{X} = \{\textrm{x}_m\}_{m=1}^N$ and $\textbf{Y} = \{\textrm{y}_m\}_{m=1}^N$ are associated to image and text modalities, respectively. $\textrm{x}_m \in \mathbb{R}^{d_i}$ and $\textrm{y}_m \in \mathbb{R}^{d_t}$ are image and text feature vectors, respectively. $d_i$ and $d_t$ denote the size of image and text feature dimensions. We assume that the training image-text pairs can be noisy, in which an unknown number of pairs are mismatched, but a small subset $\textbf{O}_C \in \textbf{O}$ of clean image-text pairs is available in the training set as $\textbf{O}_C = \{\textbf{X}_C, \textbf{Y}_C\}^{N_C}$, where $\textbf{X}_C \in \textbf{X}$, $\textbf{Y}_C \in \textbf{Y}$ and $N_C < N$. To reduce the adverse effect of the noisy correspondences, we propose CHNR that aims at learning a noise discriminator $D_N$ and two hash functions $f$ and $g$ for image and text modalities, respectively. Using the clean subset $\textbf{O}_C$, the noise discriminator $D_N$ learns to identify clean and noisy in the joint features $\textbf{Z} = \{\textrm{z}_m\}_{m=1}^N$. $D_N$ assigns a noise likelihood score (i.e., weight) to each pair $\textbf{W} = \{\textrm{w}_m\}_{m=1}^N$, where, $\textrm{z}_m = concat(\textrm{x}_m, \textrm{y}_m)$ and $\textrm{w}_m = D_N(\textrm{x}_m, \theta_{D_N})$, for which $concat(.)$ is a vector concatenation and $\theta_{D_N}$ are parameters of $D_N$. Joint-features and corresponding weights of clean subset $\textbf{O}_C$, are denoted as $\textbf{Z}_C$ and $\textbf{W}_C$, respectively. Using the training set $\textbf{O}_C$, hash functions $f$ and $g$ learn to generate binary hash codes $\textbf{B}_i = f(\textbf{X},\theta_i)$ and $\textbf{B}_t = g(\textbf{Y},\theta_t)$, where $\textbf{B}_i \in \{0 , 1\}^{N \times B}$ and $\textbf{B}_t \in \{0 , 1\}^{N \times B}$ for image and text modalities, respectively. $\theta_i$, $\theta_t$ are parameters of image and text hashing networks and $B$ is the length of binary hash code. To learn the hash functions $f(.)$ and $g(.)$ and noise discriminator $D_N$, the proposed CHNR includes three main modules: i) the feature extraction module that produces feature representations for image and text modalities; ii) the noise detection module that aims to detect semantically incoherent feature pairs; and iii) the hashing module that generates binary representations. The block diagram of the CHNR is shown in Fig. \ref{fig:main-diagram}. The training process of the proposed CHNR is conducted in two phases: i) a meta-learning phase, where the noise discriminator and the hashing module are trained on the clean subset $\textbf{O}_C$ only; ii) the main training phase, where the weights of $D_N$ are frozen and it is used to detect noisy correspondences while the hashing module is trained on the training set $\textbf{O}$. 

\subsection{Feature extraction module}
\label{sec:method-components-feature}

This module generates deep semantic representations for both image and text modalities, and feed them into the noise detection and hashing modules. The feature extraction module includes two modality-specific encoder networks: 1) an image encoder network $net_{img}$; 2) a text (i.e., image captions) encoder network $net_{txt}$. During the training of the noise detection and the hashing modules, the weights of image and text encoders are frozen. Given the training set $\textbf{O}$, the image, text and joint-features are denoted $net_{img}(\textbf{X})$, $net_{txt}(\textbf{Y})$ and $\textbf{Z}$, respectively. For the sake of simplicity we refer $net_{img}(\textbf{X})$ as $\textbf{X}$, and $net_{txt}(\textbf{Y})$ as $\textbf{Y}$ in the rest of this paper. For the unsupervised contrastive representation learning of CHNR, we generate a corresponding augmented set from $\textbf{O}$, which is defined as $\textbf{O}^\prime = \{\textbf{X}^\prime, \textbf{Y}^\prime\}^N$, where $\textbf{X}^\prime = \{\textrm{x}_m^\prime\}_{m=1}^N$ and $\textbf{Y}^\prime = \{\textrm{y}_m^\prime\}_{m=1}^N$ are augmented image and caption where $\textrm{x}_m^\prime \in \mathbb{R}^{d_i}$ and $\textrm{y}_m^\prime \in \mathbb{R}^{d_t}$. 

The embeddings of augmented images and captions are extracted by $net_{img}$ and $net_{txt}$, respectively. For the sake of simplicity in the rest of this paper we refer $net_{img}(\textbf{X}^\prime)$ as $\textbf{X}^\prime$, and $net_{txt}(\textbf{Y}^\prime)$ as $\textbf{Y}^\prime$. Joint-features of augmented image-text pairs and corresponding weights are denoted as $\textbf{Z}^\prime$ and $\textbf{W}^\prime$ respectively. The same notation principle applies to the augmented subset without the noisy correspondences $\textbf{O}_C^\prime \in \textbf{O}^\prime$. $\textbf{X}_C^\prime$, $\textbf{Y}_C^\prime$, $\textbf{Z}_C^\prime$ and $\textbf{W}_C^\prime$ denote augmented clean image features, text features, joint-features and corresponding feature pair weights, respectively.

\subsection{Noise detection module}
\label{sec:method-components-noise}
This module aims at assigning weights to image-text pairs based on the likelihood of being noisy. The noise discriminator $D_N$ is a fully-connected network with single-neuron output that predicts if joint-feature is clean. The noise discriminator assigns low weight values to semantically incoherent (i.e., noisy) joint-features and high weights to clean pairs. During the meta-learning stage the noise discriminator is trained as a binary classifier, where original image-text feature pairs from the clean subset $\textbf{O}_C$ are concatenated in "clean" joint-features with label "1". "Noisy" joint-features with label "0" are generated with the feature mixer by randomly shuffling image and text features. Noise discriminator training loss is defined as:
\begin{align}
\label{eq:adversarial-loss}
\small
    \min_{\theta_{D_N}} \mathcal{L}_{CE}\left( \mathbb{Z}_C \right) = -\frac{1}{N_C}\sum^{N_C}\Big[ &\log\Big( D_N\big( mix(\mathbb{X}_C, \mathbb{Y}_C) \big) \Big) +\\
    &\log \Big( 1 - D_N\big( \mathbb{Z}_C \big) \Big) \Big],\nonumber
\end{align}
where $\mathbb{Z}_C = \{\textbf{Z}_C, \textbf{Z}_C^\prime\}$, $\mathbb{X}_C = \{\textbf{X}_C, \textbf{X}_C^\prime\}$, $\mathbb{Y}_C = \{\textbf{Y}_C, \textbf{Y}_C^\prime\}$ and $mix(X, Y) = concat(X, shuffle(Y))$ is the feature mixer function for the generation of semantically incoherent joint-features, where $shuffle(.)$ is a random shuffle function. During the main training phase the parameters $\theta_{D_N}$ of the noise discriminator $D_N$ are frozen. The noise discriminator $D_N$ discriminates image-text features $\textbf{Z}$ and $\textbf{Z}^\prime$ of the noisy dataset $\textbf{O}$ to generate weights $\textbf{W}$ and $\textbf{W}^\prime$, which are passed into the hashing learning module. To reduce the impact of noisy pairs we exclude them by thresholding $\textbf{W}$ and use discrete weights $\textbf{W}_D = threshold(\textbf{W})$, where $threshold(x) = \left\{\begin{matrix} 1, & x >= 0.5\\ 
0, & x < 0.5 & \end{matrix}\right.$.


\subsection{Hashing module}
\label{sec:method-components-hash}

This module aims at learning two hash functions $f$ and $g$ for cross-modal binary hash code $\textbf{B}$ generation from the image features $\textbf{X}$, $\textbf{X}^\prime$ and text features $\textbf{Y}$, $\textbf{Y}^\prime$. Joint-feature weights $\textbf{W}$, $\textbf{W}^\prime$ are generated by the noise discriminator $D_N$ and are used to reduce the impact of noisy pairs on the learning process by reducing the importance of pairs identified as noisy. The caption-noise-sensitive contrastive loss is the main objective for unsupervised representation learning in the proposed CHNR. We also employ quantization loss to improve the approximation of generated continuous binary-like values to the discrete hash code.
We use both inter-modal and intra-modal contrastive losses for better representation learning. The inter-modal term maps both modalities into a common feature space, while intra-modal terms improve the mapping within modalities. The normalized temperature scaled cross-entropy (NTXent) objective function \cite{chen2020deep} is used for contrastive losses calculation. To obtain the caption-noise-sensitive contrastive losses we introduce additional re-weighting term to reduce the impact of noisy image-text pairs on the training. The weighted inter-modal contrastive loss $\mathcal{L}_{C_{inter}}$ between image $\textrm{x}_j$ and its paired caption $\textrm{y}_j$ with image-text semantic coherence weight $\textrm{w}_j$ is computed as:
\begin{align}
\label{eq:contrastive-loss-inter}
\small
    &\mathcal{L}_{C_{inter}}(\textrm{x}_j,\textrm{y}_j) =\\
    &-\textrm{w}_j\log \frac{S\left( f(\textrm{x}_j), g(\textrm{y}_j) \right) }{ \sum_{k=1, k\neq j}^{M} S\left( f(\textrm{x}_j), f(\textrm{x}_k) \right) + \sum_{k=1}^{M} S\left( f(\textrm{x}_j), g(\textrm{y}_k) \right) }, \nonumber
\end{align}
\noindent where $S(\textrm{u},\textrm{v}) = \exp\left( \cos\left( \textrm{u},\textrm{v} \right) / \tau \right)$, and $\cos\left( \textrm{u}, \textrm{v} \right) = \textrm{u}^T\textrm{v}/\left\| \textrm{u} \right\|\left\| \textrm{v} \right\|$ is the cosine similarity, $\tau$ denotes a temperature, and $M$ is a batch size. During the meta-learning phase weight of all pairs is set to $1$ ($\textrm{w}_j = 1$), while during the main training phase the weight values are assigned by $D_N$. Image and text intra-modal contrastive losses are defined as: 

\begin{align}
\label{eq:contrastive-loss-intra-img}
\small
    &\mathcal{L}_{C_{img}}(\textrm{x}_j,\textrm{x}^\prime_j) =\\
    &-\hat{\textrm{w}}\log \frac{S\left( f(\textrm{x}_j), f(\textrm{x}^\prime_j) \right) }{ \sum_{k=1, k\neq j}^{M} S\left( f(\textrm{x}_j), f(\textrm{x}_k) \right) + \sum_{k=1}^{M} S\left( f(\textrm{x}_j),f(\textrm{x}^\prime_k) \right) }, \nonumber
\end{align}

\begin{align}
\label{eq:contrastive-loss-intra-txt}
    &\mathcal{L}_{C_{txt}}(\textrm{y}_j,\textrm{y}^\prime_j) =\\
    &-\hat{\textrm{w}}\log \frac{S\left( g(\textrm{y}_j), g(\textrm{y}^\prime_j) \right) }{ \sum_{k=1, k\neq j}^{M} S\left( g(\textrm{y}_j), g(\textrm{y}_k) \right) + \sum_{k=1}^{M} S\left( g(\textrm{y}_j),g(\textrm{y}^\prime_k) \right) }, \nonumber
\end{align}

\noindent where $\mathcal{L}_{C_{img}}$ is the contrastive loss between image $\textrm{x}_j$ and its augmented view $\textrm{x}^\prime_j$ and $\mathcal{L}_{C_{txt}}$ is the contrastive loss between caption $\textrm{y}_j$ and its augmented view $\textrm{y}^\prime_j$. During the meta-learning stage $\hat{\textrm{w}} = 1$, for the full training the averaged weight $\hat{\textrm{w}} = \frac{1}{M}\sum_{j=1}^{M}\textrm{w}_j$ is used to avoid the skew towards intra-modal objectives in the total contrastive loss $\mathcal{L}_{C}$ defined as: 
\begin{equation}
\small
    \mathcal{L}_{C} = \mathcal{L}_{C_{inter}} + \lambda_1 \mathcal{L}_{C_{img}} + \lambda_2 \mathcal{L}_{C_{txt}},
    \label{eq:contrastive-loss}
\end{equation}
\noindent where $\lambda_1$ and $\lambda_2$ are hyperparameters for image and text intra-modal contrastive losses, respectively.

The quantization loss $\mathcal{L}_{Q}$ optimizes the difference between continuous and discrete hash values and calculated as: 
\begin{equation}
\label{eq:binarization-quant}
\small
    \mathcal{L}_{Q} = \left\| \textbf{B} - \textbf{H}_i \right\|^2_F +
    \left\| \textbf{B} - \textbf{H}_i^\prime \right\|^2_F +
    \left\| \textbf{B} - \textbf{H}_t \right\|^2_F +
    \left\| \textbf{B} - \textbf{H}_t^\prime \right\|^2_F,
\end{equation}

\noindent where $\textbf{H}_i = f( \textbf{X})$, $\textbf{H}_i^\prime = f( \textbf{X}^\prime)$, $\textbf{H}_t = g( \textbf{Y})$, $\textbf{H}_t^\prime = g( \textbf{Y}^\prime)$ are binary like codes for images, augmented images, texts and augmented texts, respectively. The binary code is updated by the following rule:
\begin{equation}
\small
    \textbf{B} = sign \left( \frac{1}{2} \left( \frac{\textbf{H}_i + \textbf{H}_i^\prime}{2} + \frac{\textbf{H}_t + \textbf{H}_t^\prime}{2} \right) \right).
    \label{eq:binarization-final-binary-code}
\end{equation}

\noindent The final loss function is the weighted sum of \eqref{eq:contrastive-loss} and \eqref{eq:binarization-quant}:
\begin{equation}
\small
    \min_{\textbf{B}, \theta_i, \theta_t, \theta_D} \mathcal{L} = \mathcal{L}_C + \alpha \mathcal{L}_Q,
    \label{eq:loss-overall}
\end{equation}
where $\alpha$ is a hyperparameter for quantization loss. Finally, for the retrieval of semantically similar captions to a query image $\textrm{x}_q$, we compute the Hamming distance between $f(net_{img}(\textrm{x}_q))$ and hash codes in the archive, and the most similar $K$ captions are retrieved. Similarly, the most similar $K$ images are retrieved for a query caption $\textrm{y}_q$ with regard to the Hamming distances estimated between $g(net_{txt}(\textrm{y}_q))$ and hash codes of the images in the archive.


\begin{table}[]
\setlength{\tabcolsep}{3pt}
\centering

\caption{The mAP@20 results for image-to-text ($I \to T$) and text-to-image ($T \to I$) retrieval for the RSICD dataset when $B=64$.}
\label{tab:abl-results}
\def\arraystretch{1.1}
\begin{tabular}{c|l|cccccc}
\hline
\multirow{2}{*}{Task}& \multirow{2}{*}{Metod} & \multicolumn{6}{c}{Injected noise rate}  \\
\cline{3-8}
\cline{3-8}
& & 5\% & 10\% & 20\% & 30\% & 40\% & 50\% \\
\hline
\multirow{4}{*}{$I \to T$}
 & CHNR & \textbf{0.786} & \textbf{0.770} & \textbf{0.739} & \textbf{0.732} & \textbf{0.717} & \textbf{0.708}  \\
 & CHNR-NW & 0.778 & 0.752 & 0.730 & 0.720 & 0.668 & 0.617   \\
 & CHNR-PTC & 0.772 & 0.749 & 0.698 & 0.660 & 0.582 & 0.525  \\
 & CHNR-WNR & 0.785 & 0.767 & 0.724 & 0.677 & 0.604 & 0.475  \\
\hline
\multirow{4}{*}{$T \to I$}
 & CHNR & 0.783 & 0.782 & \textbf{0.754} & \textbf{0.746} & \textbf{0.720} & \textbf{0.718} \\
 & CHNR-NW & 0.778 & 0.767 & 0.745 & 0.737 & 0.704 & 0.664  \\
 & CHNR-PTC & 0.776 & 0.766 & 0.727 & 0.709 & 0.648 & 0.614  \\
 & CHNR-WNR & \textbf{0.784} & \textbf{0.783} & 0.751 & 0.706 & 0.647 & 0.540 \\
\hline
\end{tabular}
\end{table}

\begin{table*}[]
\setlength{\tabcolsep}{3pt}
\centering

\caption{The mAP@20 results for $I \to T$ and $T \to I$ retrieval tasks for the RSICD and the UCM datasets when $B=64$.}
\label{tab:retrieval-results}
\def\arraystretch{1.1}
\begin{tabular}{c|l|cccccc|cccccc}
\hline
\multirow{3}{*}{Task}& \multirow{3}{*}{Metod} & \multicolumn{12}{c}{Amount of the caption noise}  \\
\cline{3-14}
& & \multicolumn{6}{c}{RSICD} & \multicolumn{6}{c}{UCM} \\
\cline{3-14}
& & 5\% & 10\% & 20\% & 30\% & 40\% & 50\% & 5\% & 10\% & 20\% & 30\% & 40\% & 50\% \\
\hline
\multirow{3}{*}{$I \to T$}
 & CHNR & \textbf{0.786} & \textbf{0.770} & \textbf{0.739} & \textbf{0.732} & \textbf{0.717} & \textbf{0.708} & \textbf{0.843} & \textbf{0.832} & \textbf{0.844} & \textbf{0.823} & \textbf{0.821} & \textbf{0.796} \\
 & JDSH\cite{liu2020joint}& 0.768 & 0.737 & 0.683 & 0.609 & 0.487 & 0.381 & 0.806 & 0.775 & 0.714 & 0.620 & 0.567 & 0.396 \\
 & DJSRH\cite{su2019deep} & 0.669 & 0.661 & 0.602 & 0.559 & 0.491 & 0.397 & 0.707 & 0.688 & 0.633 & 0.586 & 0.534 & 0.426 \\
\hline
\multirow{3}{*}{$T \to I$}
 & CHNR & \textbf{0.783} & \textbf{0.782} & \textbf{0.754} & \textbf{0.746} & \textbf{0.720} & \textbf{0.718} & \textbf{0.929} & \textbf{0.912} & \textbf{0.908} & \textbf{0.891} & \textbf{0.885} & \textbf{0.849} \\
 & JDSH\cite{liu2020joint}& 0.769 & 0.754 & 0.699 & 0.639 & 0.540 & 0.415 & 0.882 & 0.860 & 0.764 & 0.699 & 0.603 & 0.442 \\
 & DJSRH\cite{su2019deep} & 0.666 & 0.647 & 0.600 & 0.547 & 0.472 & 0.373 & 0.751 & 0.742 & 0.685 & 0.621 & 0.521 & 0.468 \\
\hline
\end{tabular}
\end{table*}

\section{Experimental Results}
\label{sec:experiments}

In the experiments, we used the RSICD \cite{lu2017exploring} and the UC Merced Land Use (denoted as UCM) \cite{yang2010bag} datasets. RSICD includes $10921$ aerial images, each of which is a section of $224 \times 224$ pixels and has $5$ corresponding captions. UCM consists of $2100$ aerial images, each of which is a section of $256 \times 256$ pixels and has $5$ captions per image. For both datasets, we used only one randomly selected caption associated to each image for training. The datasets were split randomly into train, query and retrieval sets. We applied a Gaussian blur, random rotation, and center cropping for the image augmentation, while the text augmentation was performed by the rule-based replacement \cite{wei2019eda} of noun and verb tokens with semantically similar ones. The original training sets are clean but in our experiments, we fixed 20\% and 30\% of training set for RSICD and UCM, respectively as clean training set $\textbf{O}_C$, while we inject noise to the rest.

For $net_{img}$, a pre-trained ResNet architecture \cite{he2015deep} was used (the classification layer was removed) and the image feature size $d_i$ is $512$. For $net_{txt}$, a pre-trained BERT \cite{devlin2019bert} language model was used and the text feature size $d_t$ is $768$ (which was obtained by summing the last four hidden states of each token). 
For the noise discriminator $D_N$, we used a $5$-layer fully-connected network, which was trained only during the meta-learning stage. 
Hashing networks $f$ and $g$ are fully connected $3$-layer networks and a batch normalization layer after the second layer was included. 
The quantization loss hyperparameter $\alpha$ was set to $\alpha = 0.01$. Both intra-modal weights $\lambda_1$ and $\lambda_2$ from \eqref{eq:contrastive-loss} were set to $1$. The total number of training epochs was set to $150$ ($75$ for meta-training and $75$ for main training epochs). 

To analyze the effect of the noise detection module and its training strategy, we designed different configurations for the proposed CHNR as: 1) the noise detection module is not included and training is achieved on the noisy training set (denoted as CHNR-WNR); 2) the noise detection module is not included and the model initially pretrained on the small subset of the clean training set and then main training phase is applied using the noisy dataset (denoted as CHNR-PTC); and 3) the noise detection module is included but the model does not contain thresholding weights $W$ (CHNR-NW). The result of each configuration is provided in terms of mean average precision assessed on top-20 retrieved images (mAP@20) in Table \ref{tab:abl-results} in the framework of image-to-text ($I \to T$) and text-to-image ($T \to I$) retrieval tasks when $B=64$ for the RSICD dataset. The table shows that using the meta-training phase without the noise detection module can reduce the performance except for the extreme noise rate (e.g., 50\%). As an example, when the injected noise rate is 20\% CHNR-WNR provides about 3\% higher mAP@20 than CHNR-PTC for both $I \to T$ and $T \to I$ tasks. However, when the noise rate is 50\% CHNR-PTC results in more than 5\% higher mAP@20 than CHNR-WNR for both $I \to T$ and $T \to I$ tasks. Training on clean samples in the meta-learning phase prevents the model from fully learning the data distribution in the main training phase. This can reduce the performance when the noise rate is low since the model can not entirely learn from the main training phase. However, when the training set is extremely noisy, the model will not be distracted by the noise. Furthermore, the Table \ref{tab:abl-results} shows the superiority of CHNR and CHNR-NW, particularly when the noise rate is high. As an example, when the noise rate is 20\%, CHNR leads to about 5\% higher mAP@20 than CHNR-WNR for both $I \to T$ and $T \to I$. This shows that using the noise detection module makes the model robust to the noise in the training set. For smaller noise rates (i.e., 5\% and 10\%) CHNR performs almost comparable with CHNR-WNR. This is because of excluding some hard but informative samples from the training via reweighing them through the noise detection module. 

We evaluated the effectiveness of the proposed CHNR method with respect to the state of the art unsupervised cross-modal retrieval methods, which are: DJSRH \cite{su2019deep} and JDSH \cite{liu2020joint}. We trained all models under the same experimental setup for a fair comparison. Results of each method were provided in terms of mAP@20. The experiments were conducted over different injected noise rates (5\%, 10\%, 20\%, 30\%, 40\%, and 50\%) when $B=64$. Table \ref{tab:retrieval-results} shows the retrieval performance for the RSICD and UCM datasets. From the table, one can observe that the proposed CHNR method sharply outperforms all the unsupervised baselines in $I \to T$ and $T \to I$ tasks for all injected noise rates on both datasets. As an example, for $I \to T$ retrieval task for the UCM dataset, when the injected noise rate in 20\% the proposed CHNR results in about 13\% and 21\% higher mAP@20 than JDSH and DJSRH, respectively. Similarly, for the RSICD dataset in $T \to I$ retrieval task when the injected noise rate is 20\%, the proposed CHNR outperforms JDSH and DJSRH with 5\% and 15\% higher mAP@20, respectively. From the table, one can observe that all the compared methods are robust to small amounts of noise (i.e., $5\%$ and $10\%$), while the mAP@20 drops significantly for all methods by increasing the injected noise rate (more than 20\% ) for $I \to T$ and $T \to I$ tasks. However, the performance drop is considerably smaller for the proposed CHNR than the other methods. As an example, when the noise injection rate is $5\%$ for $I \to T$ task in RSICD dataset, the performance of CHNR is $2\%$ and $9\%$ higher than JDSH and DJSRH, respectively. However, when the injected noise rate increases to 50\% for the same task of the same dataset the performance of CHNR is $32\%$ and $31\%$ higher than JDSH and DJSRH, respectively.


\section{Conclusion}
\label{sec:conclusion}

In this paper, we have proposed a novel unsupervised cross-modal hashing method robust to the noisy image-text correspondences (CHNR). The proposed CHNR uses a multi-term noise-robust contrastive loss function to learn cross-modal hash codes in an unsupervised manner. In detail, the proposed loss function has weighted intra- and inter-modal objectives, which take into account the coherence of cross-modal correspondences represented by weights. For the weight calculation, the cross-modal joint-feature noise discriminator has been introduced. Furthermore, we have analyzed the proposed noise detection module and demonstrated the effectiveness of the proposed CHNR method through the experimental results. As future work, we plan to learn the augmentations in the feature level to improve the unsupervised contrastive learning process.

\section{Acknowledgment}
\label{sec:acknowledgments}

This work is funded by the European Research Council (ERC) through the ERC-2017-STG BigEarth Project under Grant 759764.

\bibliographystyle{IEEEbib}
\bibliography{refs}

\end{document}